\documentclass[10pt,conference,a4paper]{IEEEtran}
\ifCLASSINFOpdf
\else
\fi
\usepackage{graphicx}
\usepackage{algorithm2e}
\usepackage{paralist}
\usepackage{balance}
\usepackage{amsmath}
\usepackage{diagbox}
\usepackage{multirow}
\usepackage{url}

\begin{document}
%
\title{One-shot learning for acoustic identification of \\ bird species in non-stationary environments}

\author{\IEEEauthorblockN{Michelangelo Acconcjaioco and Stavros Ntalampiras}
\IEEEauthorblockA{Department of Computer Science\\
University of Milan\\
Milan, Italy, 20133\\
Email: michelangelo.acconcjaioco@studenti.unimi.it}}


%


\maketitle

\begin{abstract}
This work introduces the one-shot learning paradigm in the computational bioacoustics domain. Even though, most of the related literature assumes availability of data characterizing the entire class dictionary of the problem at hand, that is rarely true as a habitat's species composition 
is only known up to a certain extent. Thus, the problem needs to be addressed by methodologies able to cope with non-stationarity. To this end, we propose a framework able to detect changes in the class dictionary and incorporate new classes on the fly. We design an one-shot learning architecture composed of a Siamese Neural Network operating in the logMel spectrogram space. We extensively examine the proposed approach on two datasets of various bird species using suitable figures of merit. Interestingly, such a learning scheme exhibits state of the art
performance, while taking into account extreme non-stationarity cases.
\end{abstract}


%
\IEEEpeerreviewmaketitle

\section{Introduction}
Computational bioacoustics comprises a relatively recent scientific field placed on the crossroad of several disciplines, such as biology, computer science, etc. \cite{Stowell2017,Ntalampiras2018eco,Ntalampiras2019insects}. Animals use sound vocalizations as a very effective means of communication since acoustic waves 
\begin{inparaenum}[a)]
    \item can convey relevant information which cannot be transmitted in any other way,
    \item remains practically unaffected by lighting conditions, e.g. dense forests, night, etc.,
    \item do not necessitate visual contact between emitted and receiver, 
    \item can move over long distances without significant alterations, etc.
\end{inparaenum}


Interestingly, such vocalizations provide a source of information that can be used to explore the composition of this diversity in particular regions of interest. Acoustic surveying lends itself to rapid assessment programs
which quickly assess the biodiversity of specific regions \cite{Riede1996}. This translates into not only higher species counts, but also faster estimations of biodiversity \cite{Riede1998}. The work presented in \cite{10.1093/auk/108.2.443} describes how in 7 days the author recorded 
the vocalizations of 85\% of the 287 species of avifauna his team of 7 ornithologists inventoried after 54 days of intensive field work within a 2 $km^2$ area in Amazonian Bolivia, which required 36,804 mist-net hours.	In addition, such a research path is strongly motivated by major environmental challenges including invasive species, infectious diseases, climate and land-use change, etc. where automated monitoring of animals' populations can provide important information, such as
\begin{inparaenum}[a)]
    \item monitoring of range shifts of animal species due to climate change, 
    \item biodiversity assessment and inventorying of an area, c) estimation of
    \item species richness and abundance, and 
    \item assessing the status of threatened species.
\end{inparaenum}

When combined with the present massive availability of automated recording units (ARUs)\footnote{\url{https://www.wildlifeacoustics.com/products/song-meter-sm4}} it becomes evident why remote, systematic and non-intrusive, acoustic biodiversity surveys are gaining popularity in the last decade \cite{Wolfgang2016}. Such technology can assist towards complete bio-inventories of the study site and generate data about biodiversity composition within groups of taxa at multiple levels \cite{Grill2017_eusipco,Williams2010}. Acoustic monitoring can provide baseline information about specific groups of acoustically active biota, and to generate an index of biodiversity based on the complexity of  calls recorded within a region \cite{8259800, 7952907}. In this work the taxonomic group of interest is birds; nonetheless the proposed methodology is easily extensible to other groups (e.g. stridulating insects, anurans, bats and terrestrial animals) as long as the respective data becomes available.

\begin{figure*}[t]
	\centering
	\includegraphics[scale=0.45, trim = 0mm 10mm 0mm 0mm]{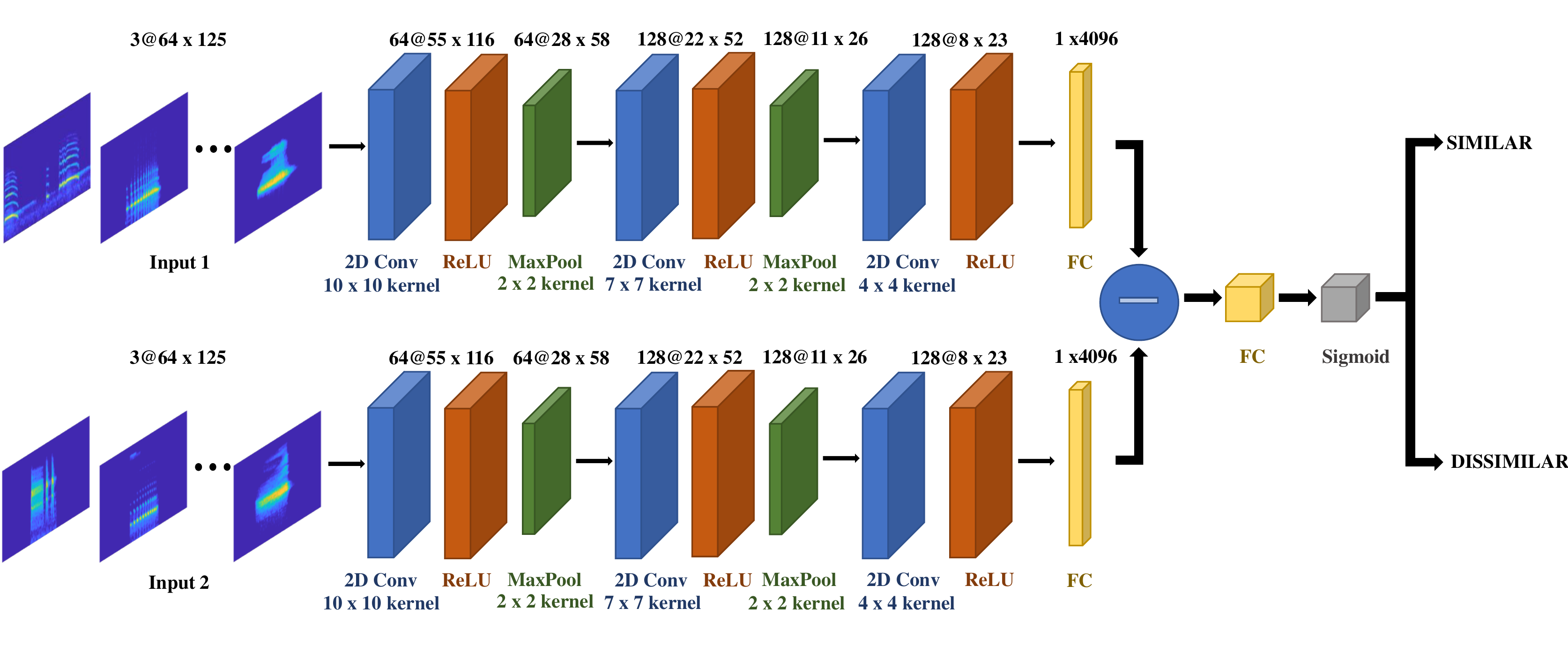}
	\caption{The pipeline of the proposed one-shot learning scheme using Siamese neural networks.}
	\label{fig:SiameseNet}
\end{figure*}

Unfortunately, there are several obstacles that need tackling towards efficient acoustic monitoring in the wild such as a potentially large and \textit{a-priori} unknown number of species \cite{Stowell2014}, big acoustic data \cite{Aide2013}, and operation under adverse conditions with non-stationary environmental noise \cite{Ntalampiras2019farm}. To the best of our knowledge, there is no solution present in the literature able to overcome these problems; classification methods able to consider unknown classes in their dictionary assume availability of a substantial amount of data in case a novel class appears \cite{DBLP:journals/corr/abs-1810-10274,7738905}, while the presence of previously unknown species, to the best of our knowledge, has not been considered in the literature. In other words, currently there are no  techniques able to perform acoustic novelty detection and bird species identification when a very small amount of data is available. To this end, we propose to explore the one-shot learning paradigm in the computational bioacoustics domain.

\textit{One-shot learning} is defined as the classification task strictly bounded by the condition that we may observe only a single sample belonging to each class in order to make inference(s) regarding test samples. In essence, the problem is solved by training a mechanism able to make predictions on the similarity of the test samples to those \textit{a-priori} available. Such a line of thought has been explored in image recognition (e.g. handwritten character recognition \cite{Koch2015SiameseNN} reaching state of the art results. In the audio signal processing domain, one-shot learning is still unexplored. 

Given the above mentioned requirements of bird species identification in the wild, this work proposes to suitably adapt the one-shot learning paradigm and having it operate on log-Mel spectrograms representing bird species. We demonstrate the efficacy of such a solution via exhaustive experiments on two datasets including the case of non-stationary environments. The main novel points of this work are:
\begin{itemize}
    \item removes the need of handcrafted features, 
    \item reaches state of the art accuracy with a very small amount of training data, and
    \item designs a reliable mechanism to detect and react to changes in the environment.
\end{itemize}

In the following, we 
\begin{inparaenum}[a)]
\item formalize the problem,
\item delineate the proposed solution,
\item describe the experimental protocol along with a detailed analysis of the obtained results, and
\item draw conclusions and briefly discuss potential extensions.
\end{inparaenum}

\section{Problem formulation}
\label{sec:problem_formulation}
This work assumes availability of a single channel audio signal $y$ containing bird vocalizations characterizing various species. We further assume that at each time instance, there is one dominating vocalization leaving the problem of simultaneous ones to future work \cite{8659544}. Composition and size of species dictionary $\mathcal{S}$ are known only up to a certain extent, meaning that new species can appear at any point in time (unknown). In other words, dictionary $\mathcal{S}=\{S_1,\dots,S_m\}$ is not bounded, while $m$ denotes the number of  species known during training. In addition, we assume that vocalizations associated with a specific species follow a consistent, yet unknown probability density function $P_i, i\in[1,m]$, as typically done in speech and generalized audio recognition algorithms \cite{Ntalampiras2019jaes,8667641}.

Without loss of generality, we assume availability of an initial training dataset $TD=y_t,t\in[1,T_0]$ with segmented labelled pairs $(y_t,S_i)$, where $t$ is the time instant and $i\in[1,m]$. No assumptions are made as to if/when a new species might appear. The overall goal is to automatically identify bird species and properly adjust $\mathcal{S}$ to deal with new ones.

\section{One-shot learning for change detection, dictionary learning and species identification}
\label{sec:oneShot}
In the present work, we employ Siamese Neural Network (SNN) for learning \textit{similar} and \textit{dissimilar} relationships trained on the data available in $TD$. Even thought this approach fits very well specifications of the current task, it is directly extensible to other classification tasks, including in non-stationary environments, with data belonging to the same (e.g. acoustic), different (e.g. image) or even multiple modalities (e.g. acoustic and image).

\subsection{Siamese Neural Networks}

SNNs originate from the work presented in \cite{NIPS1993_769} with application onto the signature verification problem. SNNs are composed by a twin network attached to a common end, albeit elaborating on different inputs as shown in Fig. \ref{fig:SiameseNet}. The shared end calculates a specific metric using the highest-level representation as extracted by each network. Since the networks 'work' towards the same goal and the optimization function is shared, their weights are tied, ensuring that similar inputs will be mapped to nearby locations in the feature space. At the same time, their topologies are symmetric, thus two different inputs to the SNN, will result to the same metric even if the networks' position (top/bottom) were to change.

Several metric functions between the twin feature vectors have been used in the literature, e.g. contrastive energy \cite{chopra2005learning}, weighted L1 distance \cite{Koch2015SiameseNN}. In this work, we employed binary cross entropy loss followed by a sigmoid activation, conveniently normalizing the output in [0,1].

\begin{figure*}[t]
	\centering
	\includegraphics[scale=0.45, trim = 20mm 0mm 0mm 0mm]{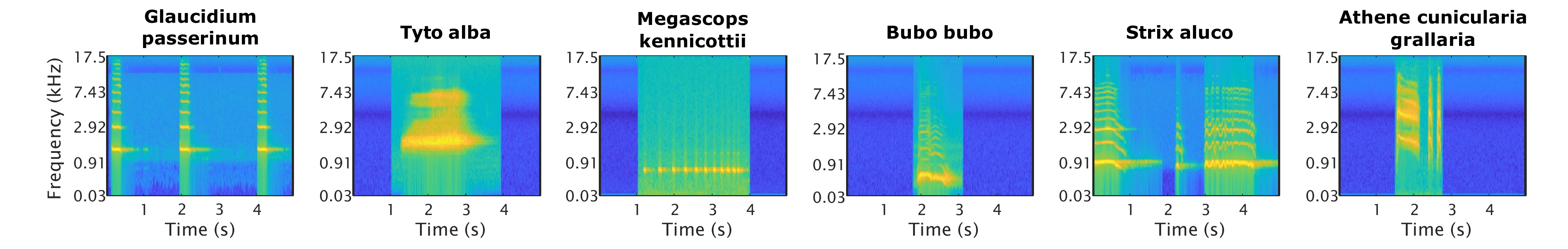}
	\caption{Log-Mel spectrograms extracted from samples belonging to all  species present in $D1$.}
	\label{fig:nocurnalSpecies}
\end{figure*}

For the present task, neural networks are composed of convolutional layers given their recent success in the field of audio signal processing \cite{8678825,7324337}. Convolutional neural networks (CNNs) are receiving increased attention in the recent years due to their simplicity and efficacy in audio classification tasks \cite{7952265}. CNNs include simple alterations in the traditional multilayer perceptron model, i.e.
\begin{inparaenum}[a)]
	\item their topology includes several stacked layers, while
	\item each convolutional layer is succeeded by a max-pooling one.
\end{inparaenum}
Importantly, stacked convolutional layers are able to arrange the neurons so that local structures of the input are highlighted in the two dimensional plane. To this end, each hidden unit is not connected to the entire input but only to a small part of it, the \textit{receptive field}. The weights associated to the hidden units (learnable convolutional kernels) are able to extract a map of features characterizing the input. Subsequently, max-pooling operations are employed in order to reduce the dimensionality of the feature maps by keeping the maximum value of neighboring units and improving the robustness of the network to translational shifts \cite{7324337}. The activation function is $f(x)=max(0,x)$, i.e. the network is composed by rectified linear units (ReLu).

\subsection{Model structure and learning}
We used two different model structures, while Fig. \ref{fig:SiameseNet} demonstrates the three-layered one. The structure includes $N$ convolutional layers, where $N=\{3,4\}$, each one followed by a ReLu and a max-pooling layer, except the last one where max-pool is substituted by a fully connected one. The SNN is completed by a distance operation, a fully connected layer and a sigmoid function responsible to decide on the inputs' affinity (similar/dissimilar) via thresholding its output.

\begin{algorithm}[t]
	1. Input: test vocalization $y^t$, trained SNN $\mathcal{N}$, dictionary $\mathcal{S}=\{S_1,\dots,S_m\}$, while each class is represented by extracted log-Mel spectrograms $\langle \mathcal{F}_\mathcal{S}^i \rangle_{i=1}^{i=|S|}$\;
	2. Extract log-Mel spectrogram $logMel$ of $y^t$ \;
	3. Initialize similarity vector $V=[]$\;
	4. \For{j=1:m}{
		5. \For{i=1:$|S|$}{
			6. Query $\mathcal{N}$ with the pair $\{logMel,\mathcal{F}_j^i\}$ and get similarity score $V(j,i)$\; 
		}
		}
	7. Predict the class maximizing the similarity score $S^*=
\underset{\mathcal{S}}{\operatorname{arg\,max}}\{V(:,i)\}$ \;
    8. Assign $S^*$ to $y^t$ \;
	\
	\caption{The proposed bird species identification algorithm based on one-shot learning ($|\bullet|$ denotes the cardinality operator).}
	\label{alg:oneTesting}
\end{algorithm}

The convolutional layers encompass filters of varying size with a constant stride equal to 1. The standard ReLU activation function is used, while max-pool layers have \(2\times2\) kernels with \(stride = 2\) . Subsequently, a flattening layer collects all units of the last convolutional twin layers and the distance calculation follows. In Fig. \ref{fig:SiameseNet}, \(N = 3\).

The binary cross-entropy loss between the network prediction and the true label is used to update the network during training. At the same time, the standard backpropagation algorithm was used with the gradient summing the weights of each twin network. The minibatch size depends on the size of the training set and the learning rate is $6\mathrm{e}{-5}$. Weights and biases were initialized using narrow normal distributions with zero-mean and 0.01 standard deviation. 

\subsection{Log-Mel spectrogram} 

Towards minimizing the need of handcrafted features, we simply extract log-Mel spectrograms of the available vocalizations to feed the SNN. In brief, we followed the traditional pipeline based on the short time Fourier transform. After early experimentations, we used 128 equal-width log-energies while their overlapping is dictated by the Mel filter bank.

\subsection{Change detection and species identification}

The SNN depicted in Fig. \ref{fig:SiameseNet} is trained with the objective of learning to identify similar and dissimilar pairs of input log-Mel spectrograms. As such, we propose a straightforward extension of one-shot learning to address change detection tasks. A change is signaled when a new log-Mel spectrogram is predicted as dissimilar with respect to all sound classes in dictionary $\mathcal{S}$. In such a case, a new class is formed and populated by the new log-Mel spectrogram alone. At the same time, dictionary $\mathcal{S}$ is suitably updated. On the contrary, the class with the highest similarity log-Mel spectrograms is assigned to the new sample. Conveniently, SNN has the possibility to address classification tasks with underpopulated classes effectively \cite{Koch2015SiameseNN}.

\begin{figure*}[t]
	\centering
	\includegraphics[scale=0.45, trim = 20mm 0mm 0mm 0mm]{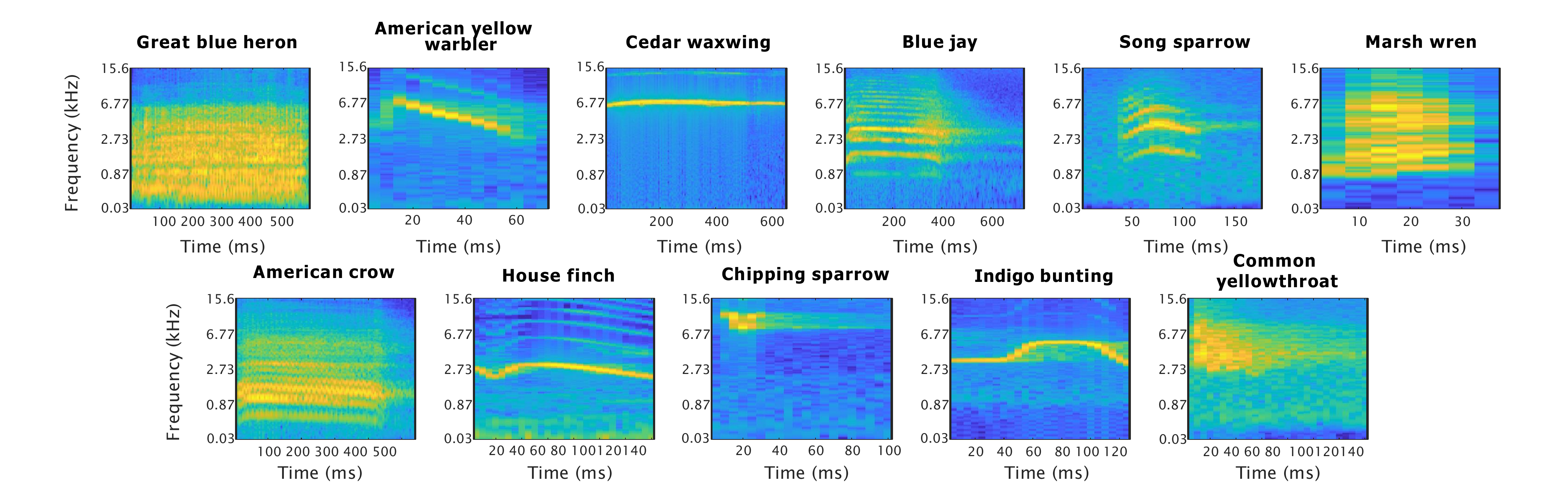}
	\caption{Log-Mel spectrograms extracted from samples belonging to all  species available in $D2$.}
	\label{fig:NAspecies}
\end{figure*}

In case a change is not detected, species identification is carried out as outlined in Alg. \ref{alg:oneTesting}. The algorithm needs four inputs (Alg. \ref{alg:oneTesting}, line 1)
\begin{itemize}
\item a test vocalization $y^t$, 
\item the trained SNN $\mathcal{N}$, 
\item the dictionary $\mathcal{S}=\{S_1,\dots,S_m\}$, while each class is represented by extracted log-Mel spectrograms $\langle \mathcal{F}_\mathcal{S}^i \rangle_{i=1}^{i=|S|}$, and
\item the available log-Mel spectrograms $\langle \mathcal{F}_\mathcal{S}^i \rangle_{i=1}^{i=|S|}$.
\end{itemize}
After extracting the log-Mel spectrogram $logMel$ of $y^t$ (Alg. \ref{alg:oneTesting}, line 2) and initializing the similarity vector $V$ (Alg. \ref{alg:oneTesting}, line 3), we query $\mathcal{N}$ with all available pair combinations and store the obtained similarity scores in $V$ (Alg. \ref{alg:oneTesting}, line 4-6). Finally, the class maximizing the similarity score is predicted as the label of $y^t$ (Alg. \ref{alg:oneTesting}, line 7).

\begin{table}[b]
\centering
\caption{$k$-NN, SVM, 3ConvSNN and 4ConvSNN average recognition rates (in \%) on dataset $D1$. The highest rate for each percentage split is emboldened.}
\label{tab:TabD1}
\begin{tabular}{|c|c|c|c|c|c|} 
\hline
\diagbox{\textbf{Method}}{\textbf{{$split$}}}
    & \textbf{10\%} & \textbf{20\%} & \textbf{30\%} & \textbf{50\%} & \textbf{70\%}  \\ 
\hline
\hline
\textbf{\textit{k}-NN }    & \textbf{89.26}         & \textbf{93.23}         & 92.17         & \textbf{93.9}         & 95.83          \\ 
\hline
\textbf{SVM}   & 81.88~        & 87.22         & 89.57         & 92.68         & 93.75          \\ 
\hline
\multicolumn{6}{|c|}{ \textbf{3ConvSNN} }                                               \\ 
\hline
\textbf{Mean}       & 75.01         & 84.38         & 92.01         & 91.92         & 96.03          \\ 
\hline
\textbf{Std}  & 1.39          & 1.12          & 1.36          & 1.05          & 1.27           \\ 
\hline
\multicolumn{6}{|c|}{\textbf{4ConvSNN} }                                               \\ 
\hline
\textbf{Mean}       & 75.49         & 86.63         & \textbf{93.3}          & 92.94         & \textbf{97.44}          \\ 
\hline
\textbf{Std}  & 1.52          & 1.68          & 1.28          & 1.28          & 1.20           \\
\hline
\end{tabular}
\end{table}

\section{Experimental set-up and results}
This section describes the 
\begin{inparaenum}[a)]
	\item employed datasets,
	\item suitably formed figures of merits, 
	\item contrasted methods, and 
	\item obtained results.
\end{inparaenum}

\subsection{Datasets}

In order assess extensively the performance achieved by the proposed method we used two datasets:

\subsubsection{$D1${\label{item:D1}}} is focused on nocturnal bird species, a task which is rather new for the computational bioacoustics community. Despite their significance in terms of migrations trends and patterns, when considered, nocturnal bird species comprise a relatively small part of the dataset \cite{Salamon2016}. Interestingly, the associated vocalizations tend be to rather clean as less environmental and/or man-made noise is typically present during night, thus the pattern recognition mechanism can operate on the vocalization structure alone.

\begin{table}[b]
\centering
\caption{Confusion matrix (in \%) obtained by the 4 Convolutional Layer SNN considering 30\% of training data of dataset $D1$.}
\label{tab:Tab4D1}
\begin{tabular}{|c|c|c|} 
\hline
\diagbox{\textbf{Presented}}{\textbf{Predicted}} & \textit{Similar} & \textit{Dissimilar}  \\ 
\hline
\textit{Similar}                                                &\textbf{ 90.9}             & 9.1                  \\ 
\hline
\textit{Dissimilar}                                             & 4.5              & \textbf{95.5}                 \\
\hline
\end{tabular}
\end{table}

$D1$ includes nocturne rapacious species. The following six were chosen covering both European and American autochthonous species: \textit{Bubo bubo} (Eagle Owl, Europe), \textit{Strix aluco} (Tawny Owl, Europe), \textit{Glaucidium passerinum} (Owl, Europe), \textit{Tyto alba} (Barn Owl, Europe), \textit{Athene cunicularia} (Burrowing Owl, South America), and  \textit{Megascops kennicottii} (Western Screech-Owl, North America). The data were obtained from the Xeno-Canto repository\footnote{Xeno Canto, \url{https://www.xeno-canto.org/about/xeno-canto}}. Log-Mel spectrograms extracted from samples belonging to all species in $D1$ are shown in Fig. \ref{fig:nocurnalSpecies}.

The dataset is composed of 167 files while each species is represented by 25-30 files maintaining a satisfactory balance. DC-offset was removed to compensate for potential microphone calibration problems. Each audio file contains 1 to 3 calls, its duration is 5s and is sampled at 44.1kHz. The frame size is 0.03s and overlap 0.015s.

\subsubsection{$D2$\label{item:D2}}
This dataset \cite{8462156} represents real-world conditions as it contains field recordings of eleven bird species, with the total number of bird vocalizations being 2762. Interestingly, several sources of interference are present, such as non-stationary noise (wind, rain, etc), sounds related to man-made activities, etc. The dataset includes recordings from the Xeno-canto archive as well\footnote{\url{https://zenodo.org/record/1250690#.XvnsjigzY2w}}.

The following species are present: \textit{Blue Jay}, \textit{Song Sparrow}, \textit{Marsh Wren}, \textit{Common Yellowthroat}, \textit{Chipping Sparrow}, \textit{American Yellow Warbler}, \textit{Great Blue Heron}, \textit{American Crow}, \textit{Cedar Waxwing}, \textit{House Finch}, \textit{Indigo Bunting}. The dataset is sampled at 32kHz and is well-balanced across species. Here, the frame size is 0.01s and overlap 0.005s. $D2$ was formed in the paper presented in \cite{8462156} along with an experimental protocol conveniently allowing reliable comparison among competing methods. Log-Mel spectrograms extracted from samples belonging to all species in $D2$ are shown in Fig. \ref{fig:NAspecies}.

\subsection{Figures of merit}
\label{sub:B}
We employed effective and widely-used figures of merit assessing the performance of all methods thoroughly. One interesting detail for the case of one-shot learning is that we can additionally employ confusion matrices at the entire dataset level (i.e. for $D1$ and $D2$) demonstrating the efficacy of the method in recognizing similarities and dissimilarities. To this end, the following matrix was defined:

\begin{equation}
\mathcal{M}^s=
\begin{bmatrix} s_{11} & s_{12} \\ s_{21} & s_{22} \end{bmatrix},
\end{equation}
where 
\begin{itemize}
    \item $s_{11}$ (in \%) denotes the number of times that samples fed in the first input of SNN were identified as similar to samples coming from the same class,
    \item $s_{12}$ (in \%) denotes the number of times that samples fed in the first input of SNN were identified as dissimilar to samples coming from the same class,
    \item $s_{22}$ (in \%) denotes the number of times that samples fed in the second input of SNN were identified as similar to samples coming from the same class,
    \item $s_{21}$ (in \%) denotes the number of times that samples fed in the second input of SNN were identified as dissimilar to samples coming from the same class.
\end{itemize}
In this case, the objective it to maximize the values in the diagonal. A matrix assessing the dissimilarities $\mathcal{M}^d$ can be defined in an analogous way with the difference being that we are aiming at minimizing its diagonal. Interestingly, the sum of similarity and dissimilarity matrices characterizing the accuracy of a given method is 100\%, i.e. $\mathcal{M}^s+\mathcal{M}^d=100$ for all elements.

\begin{table}[t]
\centering
\caption{$\mathcal{M}^s$ (in \%) achieved by 4ConvSNN with employing 30\% of training data of $D1$.}
\label{tab:Tab2D1}
\scalebox{0.9}{
\begin{tabular}{|c|c|c|c|c|c|c|} 
\hline
\diagbox{\textbf{\textit{Input 1}}}{\textbf{\textit{Input 2}}} & \textit{Burr.} & \textit{Eagle} & \textit{Owl}    & \textit{Scr.} & \textit{Tawny} & \textit{Barn}  \\ 
\hline
\hline
\textit{Burrowing Owl}                                         & \textbf{ 86.94}        & 0.33               & 5.13            & 3.77                 & -                  & 21.46              \\ 
\hline
\textit{Eagle Owl}                                             & 0.33                   & \textbf{ 96.46}    & 4.32            & 8.28                 & 9.04               & -                  \\ 
\hline
Owl                                                            & 5.13                   & 4.32               & \textbf{ 80.46} & 5.29                 & 5.99               & -                  \\ 
\hline
\textit{Screech Owl}                                           & 3.77                   & 8.28               & 5.29            & \textbf{ 91.38}      & 3.96               & -                  \\ 
\hline
\textit{Tawny Owl}                                             & -                      & 9.04               & 5.99            & 3.96                 & \textbf{ 98.83}    & -                  \\ 
\hline
\textit{Barn Owl}                                              & 21.46                  & -                  & -               & -                    & -                  & \textbf{ 91.25}    \\
\hline
\end{tabular}}
\end{table}

\subsection{Contrasted methods}
Keeping in mind the problem formulation, the proposed solution has been compared to methods able to cope with limited amount of data, i.e. the standard version of $k$-NN with the Euclidean distance as a metric \cite{WittenFrankHall11,Sirs19}. We also employed a support vector machine (SVM) composed of a radial basis function (RBF) kernel as explained in \cite{Ntalampiras2015}. Both of these methods operated in the MFCC domain, i.e. after the application of the discrete cosine transform on the log-Mel spectrograms and retaining the 13 most important coefficients.

In addition, for $D2$ the literature already includes state of the art solution \cite{8462156} using a wide range of handcrafted features focused on spectral pattern and texture modeled by an SVM with an RBF kernel. Other deep learning solutions (Convolutional/Recurrent Neural Networks) were proven to be unsuitable for the specific tasks due to overfitting.

\subsection{Experimental Protocol and Network configuration under stationary conditions}

Towards understanding the impact of the amount of training data, we used splits coming from the following set $split=\{10\%,20\%,30\%,50\%,60\%,70\%\}$. Each experiment was iterated 50 times, while data were chosen randomly for each iteration. Here, we report average and standard deviation of the achieved recognition rates. Regarding the SNN settings, the number of epochs was 2000 (early stopping was employed), MiniBatchSize 50($D1$)/100($D2$), testBatch 300, and the number of tests per class assessing similarity/dissimilarity was 15. During testing, the randomly generated similar and dissimilar input pairs are balanced.

\begin{table}[t]
\centering
\caption{$k$-NN, SVM, 3ConvSNN and 4ConvSNN average recognition rates (in \%) on dataset $D2$. The highest rate for each percentage split is emboldened.}
\label{tab:Tab1D2}
\begin{tabular}{|c|c|c|c|c|c|} 
\hline
\diagbox{\textbf{Method}}{\textbf{{$split$}}}
                                & \textbf{10\%} & \textbf{30\%} & \textbf{50\%} & \textbf{60\%} & \textbf{70\%}  \\ 
\hline\hline
\textbf{\textit{k}-NN} & 80.21         & 87.82         & 90.56         & 91.72         & 92.46 \\
\hline
\textbf{handcrafted+SVM \cite{8462156} }           & -             & -             & -             & \textbf{96.7}          & -                       \\ 
\hline
\multicolumn{6}{|c|}{\textbf{3ConvSNN} }                                                        \\ 
\hline
\textbf{Mean}               & \textbf{88.61}         & 92.09         & \textbf{93.96}         & 94.92             & \textbf{96.02}          \\ 
\hline
\textbf{Std }          & 0.85          & 0.37          & 0.37          &0.14              & 1.27           \\ 
\hline
\multicolumn{6}{|c|}{\textbf{4ConvSNN} }                                                        \\ 
\hline
\textbf{Mean}               & 88.12         & \textbf{92.41}         & 93.60         & 94             & 95.74          \\ 
\hline
\textbf{Std}          & 0.37          & 0.42          & 0.41          & 0.1            & 0.38           \\
\hline
\end{tabular}
\end{table}

\begin{table}[b]
\centering
\caption{Confusion matrix (in \%) obtained by the 3ConvSNN considering 60\% of training data of dataset $D2$}
\label{tab:test1CMD2}
\begin{tabular}{|c|c|c|} 
\hline
\diagbox{\textbf{Presented}}{\textbf{{Predicted}}} & \textit{Similar} & \textit{Dissimilar}  \\ 
\hline
\textit{Similar}                                                &\textbf{ 97.1}             & 2.9                  \\ 
\hline
\textit{Dissimilar}                                             & 7.3              & \textbf{92.7 }                \\
\hline
\end{tabular}
\end{table}

\begin{table*}[t]
\centering
\caption{$\mathcal{M}^s$  (in \%) achieved by 3ConvSNN employing 60\% of training data of $D2$}
\label{tab:merST1D2}
\begin{tabular}{|c|c|c|c|c|c|c|c|c|c|c|c|} 
\hline

\diagbox{\textbf{\textit{Input 1}}}{\textbf{\textit{Input 2}}} & \textit{ A-C}   & \textit{A-Y-W}  & \textit{ B-J}   & \textit{ C-W}   & \textit{ C-S}   & \textit{C-Y}    & \textit{G-B-H}  & \textit{ H-F}   & \textit{I-B}    & \textit{ M-W}   & \textit{ S-S}    \\ 
\hline
\hline
\textit{A-C}                                                   &\textbf{95.31} &-               &-               &-               &-               &-               &7.25            &-               &-               &-               &1                \\ 
\hline
\textit{A-Y-W}                                                 & -               &\textbf{95.04} &-               &-               &-               &41.86           &-               &-               &5.88            &4.71            &-                \\ 
\hline
\textit{B-J}                                                   &-               &-               &\textbf{93.24} & -               &1.14            &-               &1.72            &2.53            &-               &2.11            &1.32             \\ 
\hline
\textit{C-W}                                                   &-               &-               &-               & \textbf{93.43} &11.65           &-               &-               &-               &2.67            &-               &-                \\ 
\hline
\textit{C-S}                                                   &-               &-               & 1.14            & 11.65           &\textbf{88.18} & 4.48            &-               &-               &-               &4.12            &-                \\ 
\hline
\textit{C-Y}                                                   &-               &41.86           &-               &-               &4.48            &\textbf{93.85} &-               &-               &1.87            &16.3           &-                \\ 
\hline
\textit{G-B-H}                                                 &7.25            &-               &1.72            &-               & -              &-               &\textbf{91.14} &-               &-               &-               &-                \\ 
\hline
\textit{H-F}                                                   &-               &-               &2.53            &-               &-               &-               &-               &\textbf{97.67} &30.12           &-               &-                \\ 
\hline
\textit{I-B}                                                   &-               &5.88            &-               &-               &-               &1.87            &-               &30.12           &\textbf{85.75} &-               &-                \\ 
\hline
\textit{M-W}                                                   &-               &4.71            &2.11            &2.67            &4.12            &16.3           &-               &-               &-               &\textbf{86.22} &5.26             \\ 
\hline
\textit{S-S}                                                   &1               &-               &1.32            & -               & -               &-               &-               &-               &-               &5.26            &\textbf{98.41}  \\
\hline
\end{tabular}
\end{table*}

\subsection{Performance under stationary conditions on D1}

Table \ref{tab:TabD1} shows the recognition rates achieved by $k$-NN, SVM and two SNN structures. SNNs evaluation was iterated 50 times to account for the random pair selection during testing; hence, we report mean and standard deviation values for different percentage splits. As we can see the SNN, even though it is not trained specifically for classification, it is able to provide reliable performance via assessing similarities and dissimilarities with data coming from known classes ($\mathcal{S}$). Importantly, in several percentage splits, SNN is able to surpass other classification methods. In Table \ref{tab:TabD1}, we can see that, with 30\% of training data, the 4 convolutional layer SNN starts to give better performance than SVM and \textit{k}-NN. Moreover, we observe that the best performance (97.44\%) is achieved by the 4 convolutional layer SNN trained with 70\% of training data. As expected, higher rates are reached when more training data become available.

As one-shot learning is typically used in poor data availability conditions, we show the confusion matrix showing similarities and dissimilarities w.r.t 30\% split in Table \ref{tab:Tab4D1}. In general, the network recognizes better dissimilarities (95,5\%) than similarities (90,9\%), while precision is 0.95 and recall 0.91. Finally, matrix $\mathcal{M}^s$ (explained in \ref{sub:B}) is given in Table \ref{tab:Tab2D1}, while $\mathcal{M}^d=100-\mathcal{M}^s$. There, we see that \textit{Tawny Owl}-\textit{Tawny Owl} couples are correctly classified as similar with the highest rate (98.83\%). On the contrary, \textit{Owl}-\textit{Owl} and \textit{Barn Owl}-\textit{Burrowing Owl} couples are misclassified with rates 19.54\% and 21.46\% respectively.






\subsection{Performance under stationary conditions on $D2$}
Table \ref{tab:Tab1D2} tabulates the rates obtained under stationary conditions on $D2$ for all different methods and percentage splits. Similarly to $D1$ we see that SNN achieves significant rates outperforming $k$-NN in all percentage splits. Here, the proposed method is contrasted with the one presented in \cite{8462156}, where the authors used the 60\% split. We see that the SNN with 3 convolutional layers provides $94.92\pm0.14\%$ while the contrasted method 96.7\%. Even though the proposed method is slightly worse, it should be highlighted that SNN is trained only on assessing similarities of logMel spectrograms, unlike the \cite{8462156} which employs handcrafted features and SVM trained for classification.



 

Looking into the raw confusion matrix returned by the 3 convolutional layer SNN trained on 60\% of the data (Table \ref{tab:test1CMD2}), we see that the SNN recognizes better similarities (97.1\%) than dissimilarities (92.7\%), while precision is 0.97 and recall 0.93.

To conclude this part of the experiments, the corresponding $\mathcal{M}^s$ is given in Table \ref{tab:merST1D2}. We observe that the best recognized couples are \textit{House finch - House finch} (H-F) with a rate of 97.67\% and \textit{Song sparrow-Song sparrow} (S-S) with 98.41\%. The SNN is particularly effective in identifying similarities; in fact the worst rate is \textit{Indigo bunting-Indigo bunting} (I-B, 85.75\%). On the contrary, the SNN struggles to identify dissimilarities between \textit{Indigo bunting} and House finch (I-B, H-F); indeed, it wrongly identifies 30.12\% of the input couples as similar. A similar case appeared with \textit{Common yellow throat} and \textit{American yellow warbler} (C-Y, A-Y-W), where only 58.14\% cases are classified correctly. This is caused by the similarities of the respective logMel spectrograms. Importantly, thanks to the one-shot learning classification logic, such misidentifications of dissimilarities do not lead to misclassifications.


\subsection{Experimental  Protocol  and  Network  configuration  under non-stationary conditions}

The specific experimental scenario considers an unbounded dictionary $\mathcal{S}$, i.e. the SNN is tested on data belonging to classes not available during training. Focusing on dataset $D1$ we include a superset of $D1$ in the Xeno-Canto repository, i.e. the case where no nocturnal bird species is \textit{a-priori} known, while data representing two unrelated species (\textit{Crow} and \textit{Magpie}) are available.

Regarding the SNN configuration, the number of epochs was 1200  (early  stopping  was  employed), minibatch size 50($D1$)/100($D2$), test batch 300, and the number of tests per class assessing similarity/dissimilarity was 15. During testing, the randomly generated similar and dissimilar input pairs are balanced.




We evaluated the performance of the proposed one-shot learning scheme while varying the number of unknown classes. Each experimental setting was executed 50 times, and here we present mean and standard deviation of the average recognition rates. For each iteration, the training classes were selected in a random way.

The only assumption made here is that there is availability of data belonging to at least 2 during training so that the SNN may learn both similar and dissimilar relationships. The overall number of randomly generated tested input pairs ranges is 90.000.



\subsection{Performance under non-stationary conditions on $D1$}

\begin{figure}[t]
	\centering
	\includegraphics[scale=0.43, trim = 0mm 0mm 0mm 10mm]{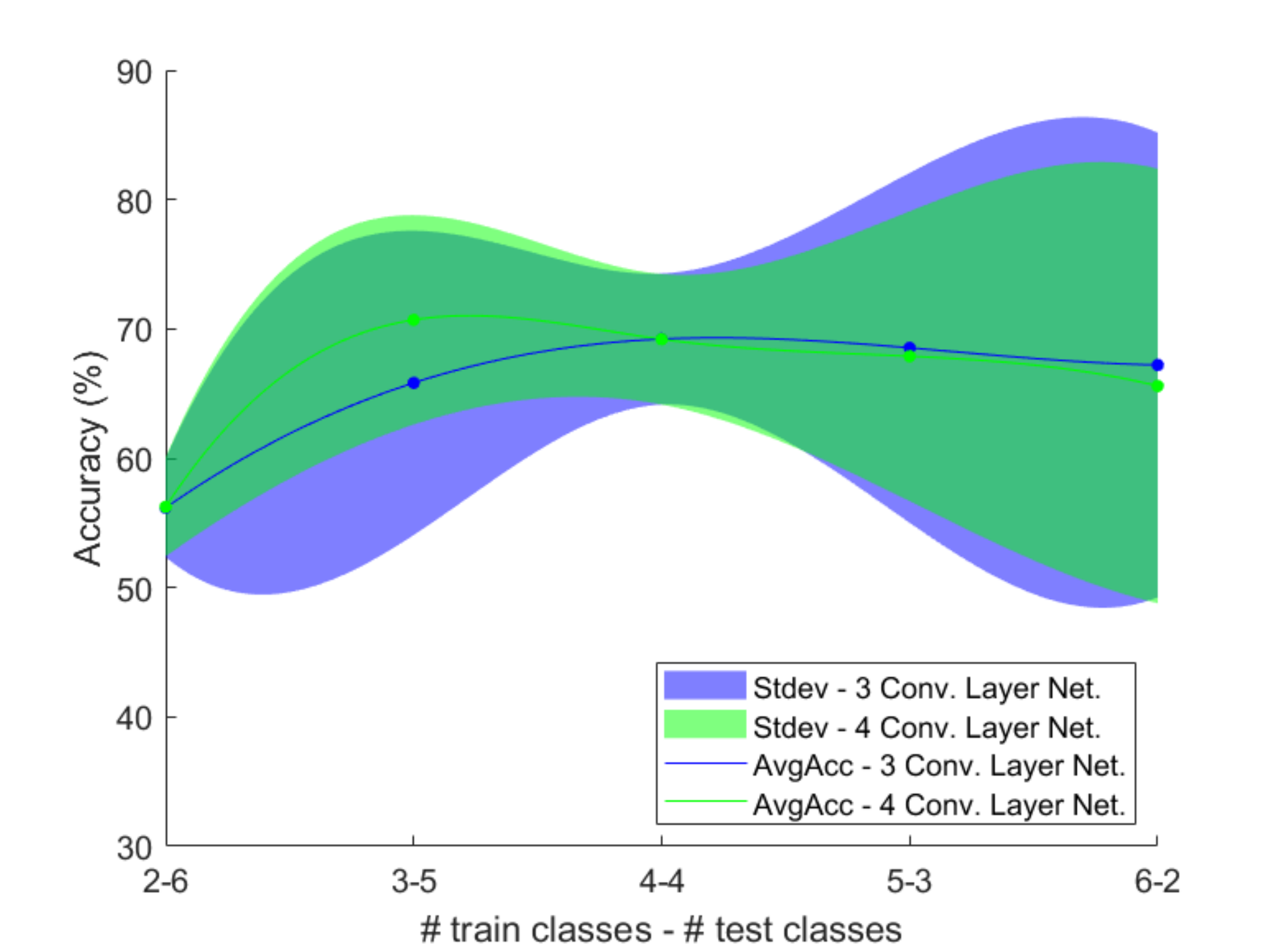}
	\caption{Average recognition accuracy (\%) in non-stationary conditions considering dataset $D1$.}
	\label{fig:Test2ChartD1}
\end{figure}

Fig. \ref{fig:Test2ChartD1} demonstrates the average recognition accuracy as a function of the number of unknown classes. In general, we observe that the rates are lower than having complete knowledge of $\mathcal{S}$, as expected. It is worth noting that average recognition rates tend to increase as more classes become available during training. The highest rate is $70.7\pm8.09\%$ reached by 4ConvSNN tested on 5 unknown classes. Overall, 4ConvSNN performs better than the 3ConvSNN. Moreover, standard deviation is lower while the networks are trained on a small number of classes. However, this depends significantly on the composition of the test and train class sets; in case  similarities/dissimilarities are easily identified, the rates increase while exhibiting lower standard deviation values. Here, class selection for both train and test sets is carried out in random way, hence the high standard deviation values.

\begin{figure}[b]
	\centering
	\includegraphics[scale=0.43, trim = 0mm 5mm 0mm 0mm]{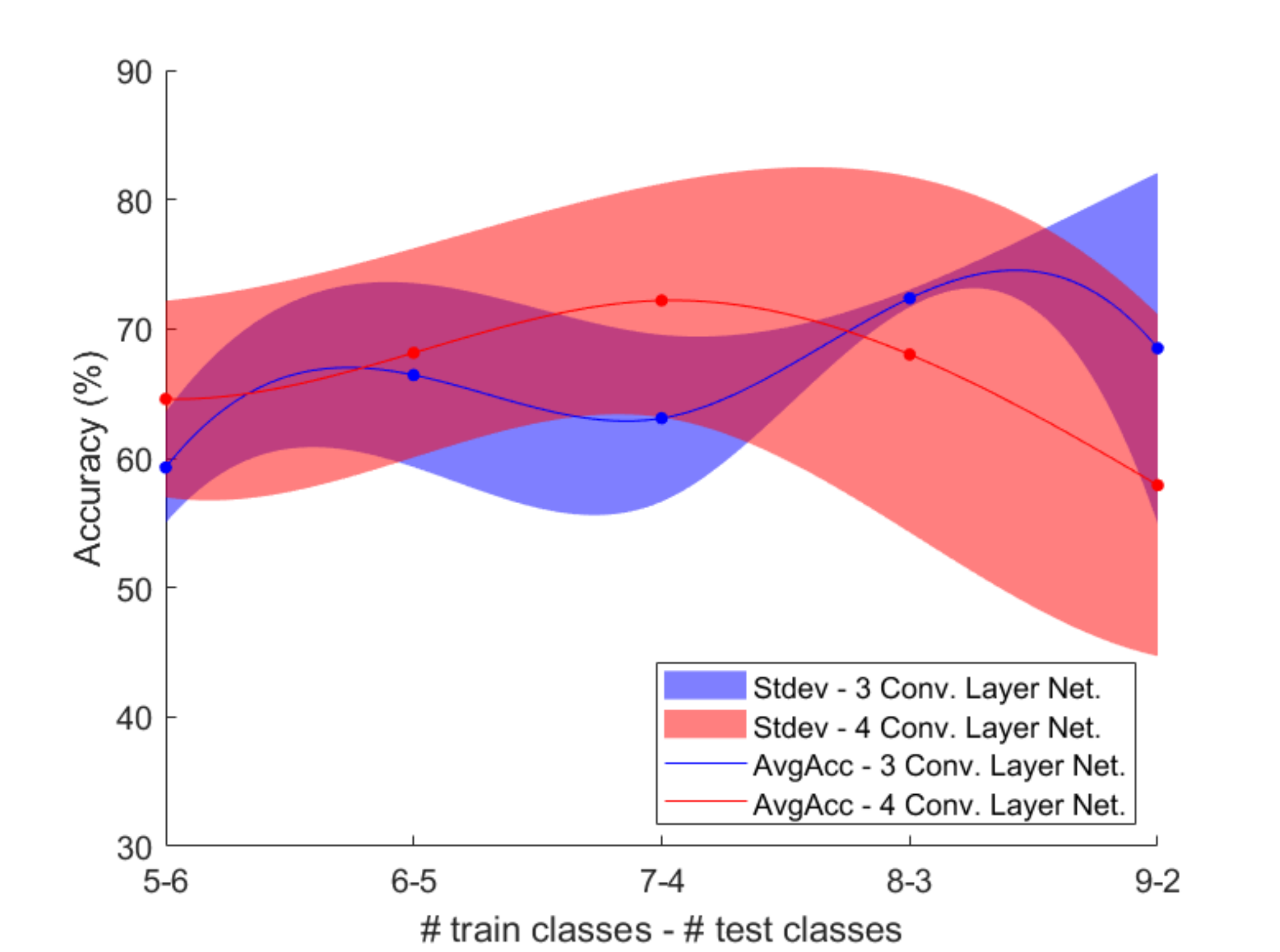}
	\caption{Average recognition accuracy (\%) in non-stationary conditions considering dataset $D2$.}
	\label{fig:Test2ChartD2}
\end{figure}

\subsection{Performance under non-stationary conditions on $D2$}
Fig. \ref{fig:Test2ChartD2} shows how the  average  recognition  accuracy alters when the number of unknown classes varies. We observe that the specific experiment reaches higher rates with lower standard deviation values w.r.t those reached on $D1$. As expected, rates are lower than those under stationary conditions while the maximum rate $72.36\pm0.67\%$ is reached by the 3ConvSNN considering 3 unknown classes. It should be noted that 4ConvSNN reaches a similar level when 4 classes are assumed to be unknown. In general, 3ConvSNN is characterized by lower standard deviation values w.r.t 4ConvSNN.

Overall, we argue that in non-stationary conditions, the performance exhibited by the one-shot learning paradigm heavily depends on the composition of the unknown class set and their similarity/dissimilarity with the classes composing the known one.


\begin{figure}[t]
	\centering
	\includegraphics[scale=0.29, trim = 0mm 5mm 0mm 0mm]{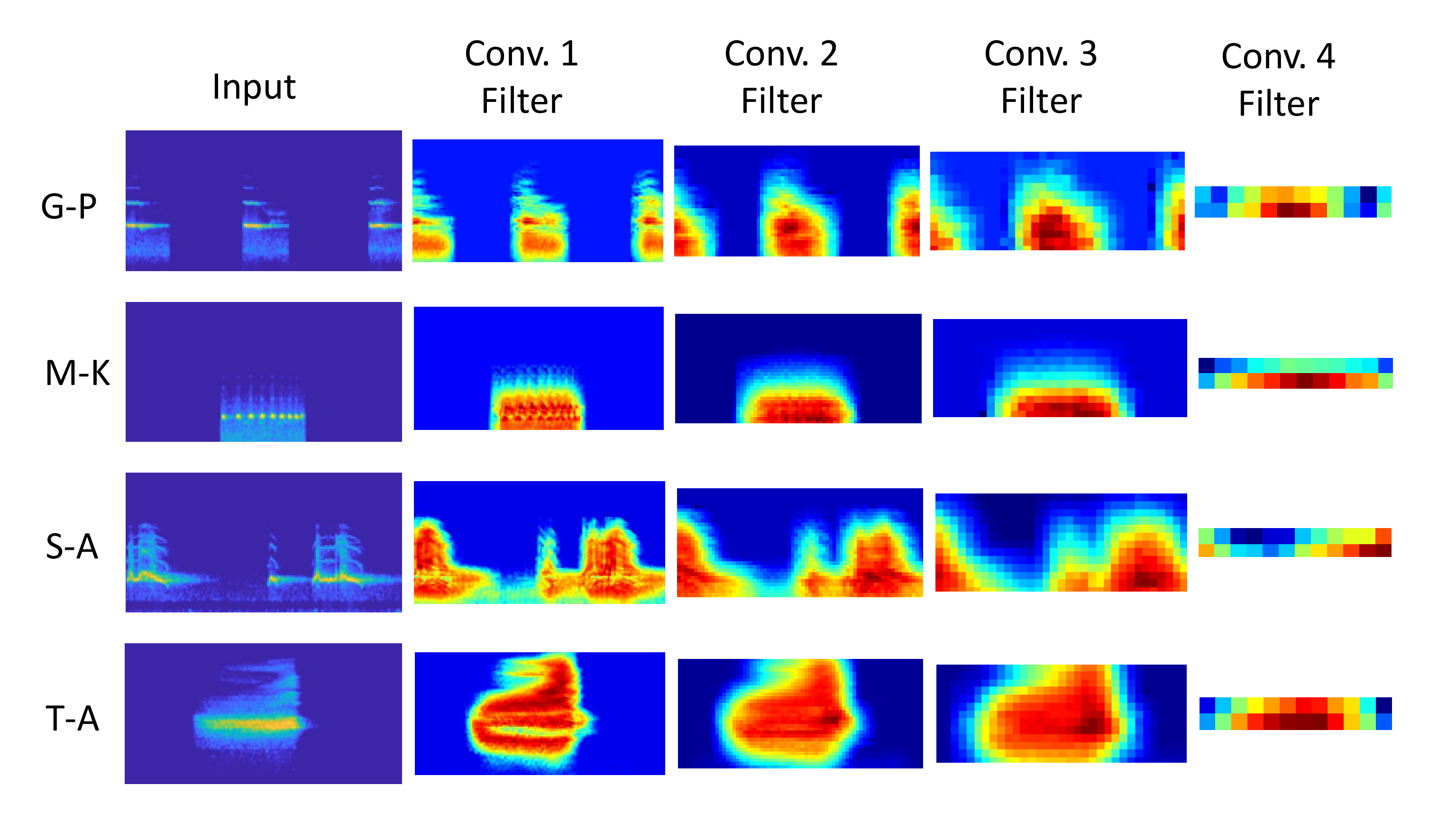}
	\caption{Convolutional layer outputs to 4 different input spectrograms taken from dataset $D1$ (4ConvSNN).}
	\label{fig:convD1}
\end{figure}

\begin{figure}[b]
	\centering
	\includegraphics[scale=0.29, trim = 0mm 0mm 0mm 10mm]{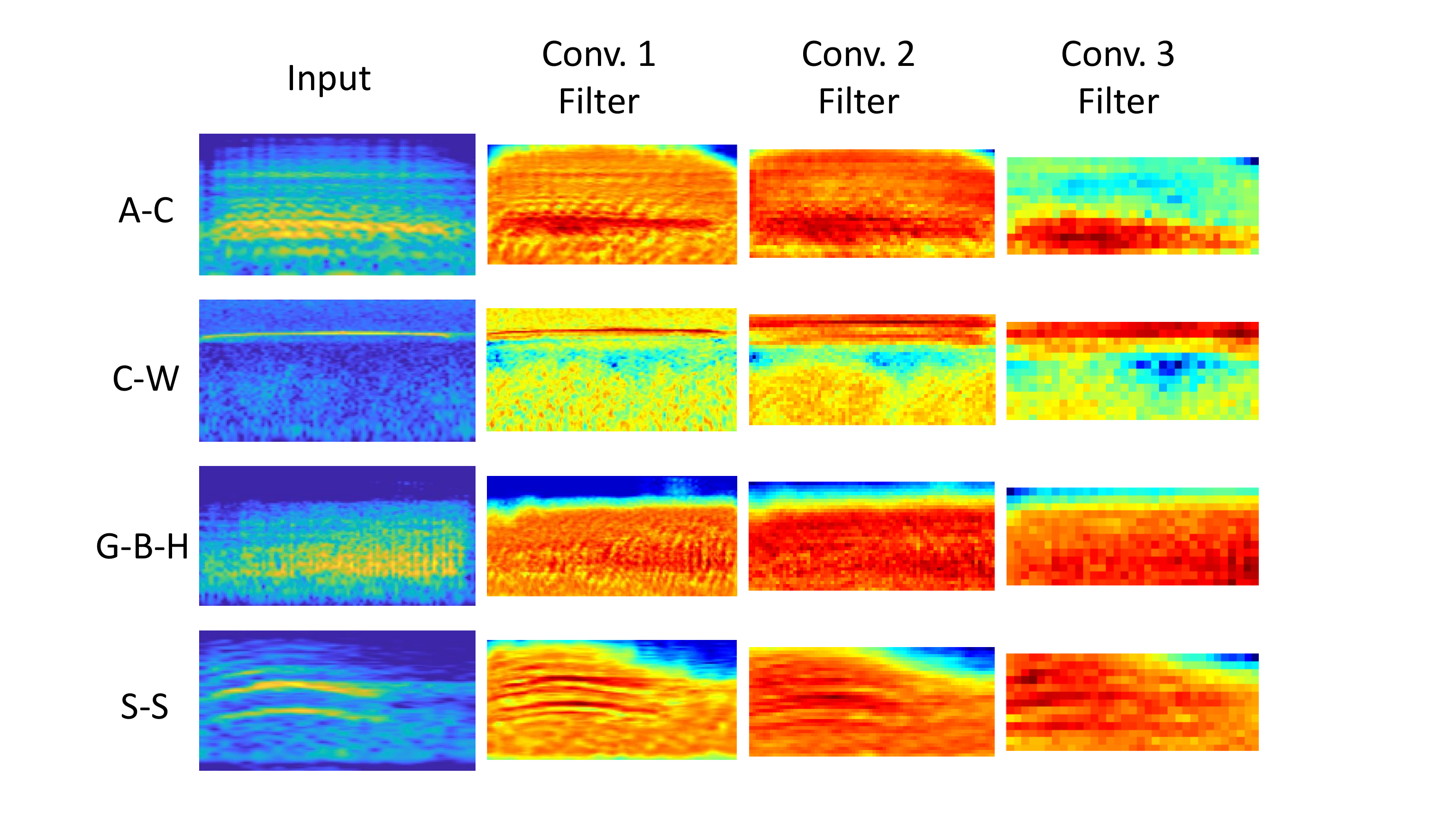}
	\caption{Convolutional layer outputs to 4 different input spectrograms taken from dataset $D2$ (3ConvSNN).}
	\label{fig:convD2}
\end{figure}

\subsection{Convolutional filters}

During the last experimental phase we examined the way the input spectrograms are processed by the network through its series of convolutional filters and localize the most significant part for the task at hand. Figures \ref{fig:convD1} and \ref{fig:convD2} demonstrate 4 different samples taken from $D1$ (\textit{Glaucidium passerinum} \textbf{G-P},  \textit{Megascops kennicottii} \textbf{M-K}, \textit{Strix aluco} \textbf{S-A}, \textit{Tyto alba} \textbf{T-A})  and $D2$ (\textit{American crow} \textbf{A-C}, \textit{Cedar waxwing} \textbf{C-W}, \textit{Great blue heron} \textbf{G-B-H},\textit{Song sparrow} \textbf{S-S}) along with their respective \textit{activation maps} for each convolutional layer of the network.

We observe that each layer simplifies the received input and focuses on the most informative region of the spectrograms. We can assert that the most distinctive feature is the distribution of the signal's energy in species-depended frequency bands. For example, when analyzing the 4th convolutional filter's outputs of \textbf{G-P} and \textbf{T-A} in Fig. \ref{fig:convD1}, we see that \textbf{T-A}'s energy is uniformly distributed in the central part of the spectrogram across all frequency bands, while \textbf{G-P}'s is primarily present in the lower bands.

\section{Conclusion}
This paper presented a method addressing the problem of acoustic bird species identification in non-stationary environments. The proposed solution, based on the one-shot learning paradigm, is able to detect changes in stationarity and incorporate unknown classes in the dictionary on the fly. Experiments carried out on two datasets showed
that the method offers state of the art performance in stationary conditions and, at the same time, it operates quite satisfactory in case of non-stationarities. Furthermore, it employs a standard audio representation eliminating the need of domain knowledge 
such as sophisticated features tailored to the problem at hand. 
We argue that a relevant part contributing to the success of this solution is its ability to consider both similarities and dissimilarities to known classes at the same time.

In the future, we wish to pursue the following directions
\begin{inparaenum}[a)]
    \item apply the present method to other problems of similar constrains examining the solution from the theoretical point of view,
    \item examine the data quantity required by the system to improve the performance in non-stationary environments, and 
    \item after verifying that class selection during training heavily influences the performance, we wish to investigate strategies enabling optimal selection of the classes composing the training set.
\end{inparaenum}


\section*{Acknowledgment}
We gratefully acknowledge the support of NVIDIA Corp. with the donation of the Titan V GPU used for this research.




\bibliographystyle{IEEEtran}
\bibliography{BiblioReconstruction}

\begin{thebibliography}{10}
\providecommand{\url}[1]{#1}
\csname url@samestyle\endcsname
\providecommand{\newblock}{\relax}
\providecommand{\bibinfo}[2]{#2}
\providecommand{\BIBentrySTDinterwordspacing}{\spaceskip=0pt\relax}
\providecommand{\BIBentryALTinterwordstretchfactor}{4}
\providecommand{\BIBentryALTinterwordspacing}{\spaceskip=\fontdimen2\font plus
\BIBentryALTinterwordstretchfactor\fontdimen3\font minus
  \fontdimen4\font\relax}
\providecommand{\BIBforeignlanguage}[2]{{%
\expandafter\ifx\csname l@#1\endcsname\relax
\typeout{** WARNING: IEEEtran.bst: No hyphenation pattern has been}%
\typeout{** loaded for the language `#1'. Using the pattern for}%
\typeout{** the default language instead.}%
\else
\language=\csname l@#1\endcsname
\fi
#2}}
\providecommand{\BIBdecl}{\relax}
\BIBdecl

\bibitem{Stowell2017}
D.~Stowell, ``Computational bioacoustic scene analysis,'' in
  \emph{Computational Analysis of Sound Scenes and Events}.\hskip 1em plus
  0.5em minus 0.4em\relax Springer International Publishing, Sep. 2017, pp.
  303--333.

\bibitem{Ntalampiras2018eco}
\BIBentryALTinterwordspacing
S.~Ntalampiras, ``Bird species identification via transfer learning from music
  genres,'' \emph{Ecological Informatics}, vol.~44, pp. 76--81, Mar. 2018.
  [Online]. Available: \url{https://doi.org/10.1016/j.ecoinf.2018.01.006}
\BIBentrySTDinterwordspacing

\bibitem{Ntalampiras2019insects}
------, ``Automatic acoustic classification of insect species based on directed
  acyclic graphs,'' \emph{The Journal of the Acoustical Society of America},
  vol. 145, no.~6, pp. EL541--EL546, Jun. 2019.

\bibitem{Riede1996}
K.~Riede, ``Diversity of sound-producing insects in a bornean lowland rain
  forest,'' in \emph{Monogr. Biologicae}.\hskip 1em plus 0.5em minus
  0.4em\relax Springer Neth., 1996, pp. 77--84.

\bibitem{Riede1998}
\BIBentryALTinterwordspacing
------, \emph{Journal of Insect Conservation}, vol.~2, no. 3/4, pp. 217--223,
  1998. [Online]. Available: \url{https://doi.org/10.1023/a:1009695813606}
\BIBentrySTDinterwordspacing

\bibitem{10.1093/auk/108.2.443}
I.~Parker, Theodore~A., ``{On the Use of Tape Recorders in Avifaunal
  Surveys},'' \emph{The Auk}, vol. 108, no.~2, pp. 443--444, 04 1991.

\bibitem{Wolfgang2016}
A.~Wolfgang and A.~Haines, ``Testing automated call-recognition software for
  winter bird vocalizations,'' \emph{Northeastern Naturalist}, vol.~23, no.~2,
  pp. 249--258, Jun. 2016.

\bibitem{Grill2017_eusipco}
T.~Grill and J.~Schl{\"u}ter, ``Two convolutional neural networks for bird
  detection in audio signals,'' in \emph{Proceedings of the 25th European
  Signal Processing Conference (EUSIPCO)}, Kos Island, Greece, Aug. 2017.

\bibitem{Williams2010}
J.~C. Williams, B.~A. Drummond, and R.~T. Buxton, ``Initial effects of the
  august 2008 volcanic eruption on breeding birds and marine mammals at
  kasatochi island, alaska,'' \emph{Arctic, Antarctic, and Alpine Research},
  vol.~42, no.~3, pp. 306--314, Aug. 2010.

\bibitem{8259800}
R.~{Kojima}, O.~{Sugiyama}, K.~{Hoshiba}, R.~{Suzuki}, and K.~{Nakadai}, ``A
  spatial-cue-based probabilistic model for bird song scene analysis,'' in
  \emph{2017 IEEE International Conference on Data Science and Advanced
  Analytics (DSAA)}, Oct 2017, pp. 395--404.

\bibitem{7952907}
Y.~{Xian}, Y.~{Pu}, Z.~{Gan}, L.~{Lu}, and A.~{Thompson}, ``Adaptive dctnet for
  audio signal classification,'' in \emph{ICASSP 2017}, Mar. 2017, pp.
  3999--4003.

\bibitem{Stowell2014}
D.~Stowell and M.~D. Plumbley, ``Automatic large-scale classification of bird
  sounds is strongly improved by unsupervised feature learning,''
  \emph{{PeerJ}}, vol.~2, p. e488, Jul. 2014.

\bibitem{Aide2013}
T.~M. Aide, C.~Corrada-Bravo, M.~Campos-Cerqueira, C.~Milan, G.~Vega, and
  R.~Alvarez, ``Real-time bioacoustics monitoring and automated species
  identification,'' \emph{{PeerJ}}, vol.~1, p. e103, Jul. 2013.

\bibitem{Ntalampiras2019farm}
S.~Ntalampiras, ``On acoustic monitoring of farm environments,'' in
  \emph{Communications in Computer and Information Science}.\hskip 1em plus
  0.5em minus 0.4em\relax Springer Singapore, 2019, pp. 53--63.

\bibitem{DBLP:journals/corr/abs-1810-10274}
J.~Pons, J.~Serr{\`{a}}, and X.~Serra, ``Training neural audio classifiers with
  few data,'' \emph{CoRR}, vol. abs/1810.10274, 2018.

\bibitem{7738905}
S.~Ntalampiras, ``Automatic analysis of audiostreams in the concept drift
  environment,'' in \emph{2016 IEEE MLSP}, Sept 2016, pp. 1--6.

\bibitem{Koch2015SiameseNN}
G.~Koch, R.~Zemel, and R.~Salakhutdinov, ``Siamese neural networks for one-shot
  image recognition,'' 2015.

\bibitem{8659544}
T.~{Ichita}, S.~{Kyochi}, and K.~{Imoto}, ``Audio source separation based on
  nonnegative matrix factorization with graph harmonic structure,'' in
  \emph{2018 Asia-Pacific Signal and Information Processing Association Annual
  Summit and Conference (APSIPA ASC)}, Nov 2018, pp. 1148--1152.

\bibitem{Ntalampiras2019jaes}
S.~Ntalampiras, ``Generalized sound recognition in reverberant environments,''
  \emph{JAES}, vol.~67, no.~10, pp. 772--781, Oct. 2019.

\bibitem{8667641}
S.~{Ntalampiras} and I.~{Potamitis}, ``A statistical inference framework for
  understanding music-related brain activity,'' \emph{IEEE Journal of Selected
  Topics in Signal Processing}, vol.~13, no.~2, pp. 275--284, May 2019.

\bibitem{NIPS1993_769}
J.~Bromley, I.~Guyon, Y.~LeCun, E.~S\"{a}ckinger, and R.~Shah, ``Signature
  verification using a "siamese" time delay neural network,'' in \emph{Advances
  in Neural Information Processing Systems 6}, J.~D. Cowan, G.~Tesauro, and
  J.~Alspector, Eds.\hskip 1em plus 0.5em minus 0.4em\relax Morgan-Kaufmann,
  1994, pp. 737--744.

\bibitem{chopra2005learning}
S.~Chopra, R.~Hadsell, and Y.~LeCun, ``Learning a similarity metric
  discriminatively, with application to face verification,'' in \emph{Computer
  Vision and Pattern Recognition, 2005. CVPR 2005. IEEE Computer Society
  Conference on}, vol.~1.\hskip 1em plus 0.5em minus 0.4em\relax IEEE, 2005,
  pp. 539--546.

\bibitem{8678825}
H.~{Purwins}, B.~{Li}, T.~{Virtanen}, J.~{Schlüter}, S.~{Chang}, and
  T.~{Sainath}, ``Deep learning for audio signal processing,'' \emph{IEEE
  Journal of Selected Topics in Signal Processing}, vol.~13, no.~2, pp.
  206--219, May 2019.

\bibitem{7324337}
K.~J. {Piczak}, ``Environmental sound classification with convolutional neural
  networks,'' in \emph{2015 IEEE 25th International Workshop on Machine
  Learning for Signal Processing (MLSP)}, Sep. 2015, pp. 1--6.

\bibitem{7952265}
S.~Mun, S.~Shon, W.~Kim, D.~K. Han, and H.~Ko, ``Deep neural network based
  learning and transferring mid-level audio features for acoustic scene
  classification,'' in \emph{ICASSP 2017}, March 2017, pp. 796--800.

\bibitem{Salamon2016}
J.~Salamon, J.~P. Bello, A.~Farnsworth, M.~Robbins, S.~Keen, H.~Klinck, and
  S.~Kelling, ``Towards the automatic classification of avian flight calls for
  bioacoustic monitoring,'' \emph{{PLOS} {ONE}}, vol.~11, no.~11, Nov. 2016.

\bibitem{8462156}
S.~{Zhang}, Z.~{Zhao}, Z.~{Xu}, K.~{Bellisario}, and B.~C. {Pijanowski},
  ``Automatic bird vocalization identification based on fusion of spectral
  pattern and texture features,'' in \emph{ICASSP 2018}, April 2018, pp.
  271--275.

\bibitem{WittenFrankHall11}
I.~H. Witten, E.~Frank, and M.~A. Hall, \emph{Data Mining: Practical Machine
  Learning Tools and Techniques}, 3rd~ed., ser. Morgan Kaufmann Series in Data
  Management Systems.\hskip 1em plus 0.5em minus 0.4em\relax Amsterdam: Morgan
  Kaufmann, 2011.

\bibitem{Sirs19}
M.~{Acconcjaioco} and S.~{Ntalampiras}, ``Acoustic identification of nocturnal
  bird species,'' in \emph{Fifth International Symposium on Signal Processing
  and Intelligent Recognition Systems (SIRS'19)}, Trivandrum, Kerala, India,
  December 2019.

\bibitem{Ntalampiras2015}
\BIBentryALTinterwordspacing
S.~Ntalampiras, ``Audio pattern recognition of baby crying sound events,''
  \emph{Journal of the Audio Engineering Society}, vol.~63, no.~5, pp.
  358--369, May 2015. [Online]. Available:
  \url{https://doi.org/10.17743/jaes.2015.0025}
\BIBentrySTDinterwordspacing

\end{thebibliography}
%

\balance

\end{document}